# An Order of Magnitude Calculus

**Nic Wilson**
Department of Computer Science
Queen Mary and Westfield College
Mile End Rd., London E1 4NS, UK
nic@dcs.qmw.ac.uk

## Abstract

This paper develops a simple calculus for order of magnitude reasoning. A semantics is given with soundness and completeness results. Order of magnitude probability functions are easily defined and turn out to be equivalent to kappa functions, which are slight generalisations of Spohn's Natural Conditional Functions. The calculus also gives rise to an order of magnitude decision theory, which can be used to justify an amended version of Pearl's decision theory for kappa functions, although the latter is weaker and less expressive.

## 1 Introduction

Order of magnitude reasoning is a common and important form of reasoning. This paper develops a simple formal calculus for such reasoning, giving a semantics for it, in terms of a new theory of infinitesimals, with soundness and completeness results. This is applied to develop order of magnitude probabilistic reasoning, in particular, an order of magnitude decision theory (partly inspired by that given in [Pearl, 93]). Such a theory might be used in an application where the information is of poor quality, so an expert may be happier just giving these very rough indications of probabilities and utilities.

Section 2 describes the mathematical construction (from [Wilson, 95]) of a new non-standard probability theory, building on ideas of Pearl [93b] and Goldszmidt [92]. Extended reals $\mathbb{R}^*$ are defined to be rational functions in parameter $\epsilon$, which is considered to be a very small unknown positive real number. Extended Probability and utility functions are defined in the usual way, except that they now can take values in $\mathbb{R}^*$. A serious problem with theories of infinitesimal probability is that it can be hard to say what these new values mean. However, section 3 (based on section 3 of [Wilson, 95]) shows how this theory overcomes the problem, giving a meaning to infinitesimal values of probability.

Section 4 introduces the order of magnitude calculus $\mathbb{R}^{\circ}$. For each integer $n$ we have an element $(+,n)$ meaning 'of order $\epsilon^n$', and an element $(-,n)$ meaning 'of order $-\epsilon^n$'. If we add something of order $\epsilon^n$ to something of order $-\epsilon^n$ then the result can be of order $\pm\epsilon^m$ for any $m \geq n$: to ensure closure of the calculus under addition we therefore add element $(0,n)$ representing this set of possibilities. The operations on $\mathbb{R}^*$ induce operations on $\mathbb{R}^{\circ}$; we explore the properties of this algebraic structure, and show how calculations can be performed within $\mathbb{R}^{\circ}$. Section 5 gives the semantics in terms of sets of extended reals, with soundness and completeness results for the calculus.

In section 6, order of magnitude probabilities and utilities are defined; the probability functions are equivalent to kappa functions, which are a slight generalisation of Spohn's Natural Conditional Functions (NCFs) [Spohn, 90], so the results of sections 3 and 5 can be used to give a formal semantics for kappa functions. An order of magnitude decision theory is constructed in section 7, and a completeness result given. The decision theory for kappa functions in [Pearl, 93a] is discussed, and partially justified by the results of this paper.

## 2 Extended Reals and Probability

This section constructs Extended Probability by first extending $\mathbb{R}$ by adding an infinitesimal element $\epsilon$, to form a new number system $\mathbb{R}^*$ and then defining probability in the usual way.

### 2.1 The Extended Reals $\mathbb{R}^*$

Let the Extended Reals $\mathbb{R}^*$ be $\mathbb{R}(\epsilon)$, the field of rational functions in (dummy variable) $\epsilon$ over the field $\mathbb{R}$ [Maclane and Birkhoff, 79, page 122]. Each ele-ment of $\mathbb{R}^*$ can be written as a pair $p/q$ where $p$ and $q$ are polynomial functions in $\epsilon$, and $p/q$ represents the same element of $\mathbb{R}^*$ as $r/s$ if and only if $ps$ is the same polynomial as $qr$. $\mathbb{R}^*$ clearly contains a copy of $\mathbb{R}$: for $x \in \mathbb{R}$, the ratio of constant polynomials $x/1$ is in $\mathbb{R}^*$, and we'll denote this element of $\mathbb{R}^*$ also by $x$. In particular the element 0 of $\mathbb{R}^*$ is the function which has constant value 0.

Every non-zero element $r$ of $\mathbb{R}^*$ can be uniquely expressed as $\bar{r}\epsilon^{\hat{r}}r'$, where $\bar{r} \in \mathbb{R} \setminus \{0\}$, $\hat{r}$ is in $Z$, the set of integers, and $r' \in \mathbb{R}^*$ is such that $r'(0) = 1$. Define $\hat{0} = \infty$. The function $r \mapsto \hat{r}$ gives the order of magnitude (in terms of powers of $\epsilon$) of element $r$ of $\mathbb{R}^*$.

### 2.2 The Ordering on $\mathbb{R}^*$

If $r = p/q \in \mathbb{R}^*$ where $p, q$ are polynomials in $\epsilon$, then, for $x \in \mathbb{R}$, $r(x)$ (the value of $r$ when $x$ is substituted for $\epsilon$) is a real number, if $q(x) \neq 0$. $\epsilon$ is intended to be a very small positive number, so we define $r > s$ iff $r$ is bigger than $s$ for small enough $\epsilon$: for $r, s \in \mathbb{R}^*$, define relation $>$ by $r > s$ if and only there exists strictly positive real number $y$ such that $r(x) > s(x)$ for all real $x$ with $0 < x < y$. Relations $<$, $\geq$ and $\leq$ are defined from relation $>$ in the usual way, e.g., $r \leq s$ if and only if $s > r$ or $s = r$.

We have $r > s$ if and only if $r - s > 0$, and (i) $r, s > 0$ implies $r + s > 0$ and $rs > 0$, and (ii) for each $r \in \mathbb{R}^*$, exactly one of the following hold: (a) $r > 0$, (b) $r = 0$, (c) $-r > 0$. Therefore $\mathbb{R}^*$ is an ordered field [Maclane and Birkhoff, 79, p261]. However, it lacks the completeness property, that every subset of positive elements has a greatest lower bound (consider, for example, the (copy of the) positive real numbers in $\mathbb{R}^*$). Thus $\mathbb{R}^*$ is not a non-standard model of the reals in the sense of Robinson's theory of hyperreals [Stroyan and Luxemburg, 76]. It is however isomorphic to a subset of the hyperreals, using the monomorphism $m$ generated by $m(x) = x$ for $x \in \mathbb{R}$ and $m(\epsilon) = \epsilon'$ where $\epsilon'$ is any positive infinitesimal.

Though they are defined as functions, elements of $\mathbb{R}^*$ should be thought of as numbers; $\epsilon$ is a positive number smaller than any strictly positive real number, $\epsilon^2$ is an even smaller positive number, $\epsilon - \epsilon^2$ is between two, though much closer to $\epsilon$, and so on.

### 2.3 Extended Probability Theory

To define Extended Probability and Utility, the usual definitions suffice, except using $\mathbb{R}^*$ instead of $\mathbb{R}$. Let $\Omega$ be a finite set, which is intended to represent a set of mutually exclusive and exhaustive events. An Extended Utility function on $\Omega$ is a function from $\Omega$ to $\mathbb{R}^*$.

Let $R$ be a set containing distinguished elements 0 and 1 with binary operation $+$ and relation $\leq$ on it, and binary operation $/$ defined on all pairs $r \in R$, $s \in R'$ where $R' \subseteq R$. Define an $R$-valued probability function P over $\Omega$ to be a function from $2^\Omega$ to $R$ satisfying

(i) $P(\emptyset) = 0$; $P(\Omega) = 1$

(ii) for $A, B \subseteq \Omega$ such that $A \cap B = \emptyset$, $P(A \cup B) = P(A) + P(B)$.

(iii) if $A \subseteq B$ then $P(A) \leq P(B)$.

For $R$-valued probability function P over $\Omega$ and $A, B \subseteq \Omega$ with $P(B) \in R'$, the conditional probability $P(A|B)$ is defined to be the value $P(A \cap B)/P(B)$.

$\mathbb{R}$-valued probability functions are just the usual probability functions (with $R' = \mathbb{R} \setminus \{0\}$). Define Extended Probability functions to be $\mathbb{R}^*$-valued probability functions (with $R' = \mathbb{R}^* \setminus \{0\}$). It can easily be checked that extended probability functions take values in $[0,1]^* = \{r \in \mathbb{R}^* : 0 \leq r \leq 1\}$.

## 3 Interpretation of Extended Probability

It is very important to be able to ascribe meaning to values of probability and utility. Clearly if an agent is going to sensibly make probability statements such as $\Pr(A) = r$, it is necessary that she understands what this means. Also, Extended Probability theory will be used to give a semantics to order of magnitude probability theory and Spohn's NCFs; the value of this semantics is heavily dependent on how strong a semantics can be given to Extended Probability theory. The issues are discussed in [Wilson, 95], and are reviewed here.

To justify the ordering used on the extended reals is fairly straight-forward. The value $\epsilon$ is considered to be an unknown small positive real number; the ordering given is the only sensible one given that we want $\mathbb{R}^*$ to be an ordered field; [Wilson, 95, 3.1] justifies the ordering axiomatically.

The axioms of Extended Probability can be justified by adapting the Dutch book argument [de Finetti, 74] or Cox's axioms [Cox, 46]; alternatively, we can view an Extended Probability function $P$ as a family of probability functions $\{\Pr_\epsilon\}$ indexed by parameter $\epsilon$ ranging over a positive neighbourhood (in $\mathbb{R}$) of 0 (this is closely related to Convergent PPDs in [Goldszmidt, 92]), and use standard justifications of (real-valued) Bayesian probability (see [Wilson, 95, 3.2]).

#### Interpretation of Probability and Utility Values

We will indicate how an agent could (in theory at least) use a sequence of thought experiments to give meaning to an assignment '$\Pr(A) = r$' for arbitrary events $A$ and for Extended Reals $r$ between 0 and 1. We will assume that an agent knows what $\Pr(A) = x$ means for real value $x \in [0,1]$, using a standard justification of Bayesian probability, either by comparison with a chance experiment (e.g. [Shafer, 81]) or a Dutch book argument (e.g., [de Finetti, 74]).[1]

The agent first imagines some (possible) event $E$ of unknown small probability, and calls this value $\varepsilon$. This step is arbitrary to quite a large extent; however once this has been chosen, the agent must stick with it for the problem at hand.

Suppose (inductively) that the agent knows what probability values of $r$ and $s$ mean where $r, s \in [0,1]^*$, and suppose (without loss of generality) that $r \leq s$. To understand the meaning of a probability of $rs$, the agent imagines two independent events $A$ and $B$ with probabilities $r$ and $s$ respectively; $A \cap B$ has probability $rs$.

If she imagines events $C$ and $D$ with $C \subseteq D$ and probabilities $r$ and $s$ respectively then the value $s - r$ is the probability of event $D \setminus C$ and the value $r/s$ is the probability of event $C$, conditional on $D$ being true. For $r, s$ with $r + s \leq 1$, the agent should imagine mutually exclusive $F$ and $G$ with probabilities $r$ and $s$ respectively; then an event should be assigned a probability value of $r + s$ if and only if it is considered equiprobable to $F \cup G$.

It turns out that any $r \in [0,1]^*$ can be reached using these rules, so by making a sequence of such thought experiments, using qualitative judgements, an agent can calibrate any of the possible values of Extended Probability.

Meaning must be given to Extended Real values of utility. We start off by assuming that an agent has, as usual, decided what 1 utile means (an arbitrary choice). Then for $r \in \mathbb{R}^*$ with $r \geq 0$, the agent imagines event $A$ with probability $\frac{r}{r+1}$ such that if $A$ occurs the agent loses 1 utile, and if $\bar{A}$ occurs the agent gains some prize $Q$. $r$ utiles is defined to be the value of a prize $Q$ which is just sufficiently large to ensure that she expects not to lose in this situation (so that the expected utility is 0). Negative values of utility are calibrated with a similar thought experiment, but where the agent gains 1 utile if $A$ occurs.

## 4 The Order of Magnitude Calculus

We will develop a calculus which can used for reasoning about the orders of magnitude of extended reals. This will be used to generate an order of magnitude theory of probability and utility.

[1] The reader may not find justifications of Bayesian probability particularly convincing, and may consider that much more general measures of belief are rational (e.g., [Walley, 91; Shafer, 81]). I would agree, but we obviously cannot hope to give a stronger justification of Extended Probability than there is of Bayesian probability;

### 4.1 Orders of Magnitude and their Meanings

Let $\mathbb{R}^o = \{(\sigma, n) : n \in Z, \sigma \in \{+,-,0\}\} \cup \{(0,\infty)\}$, where $Z$ is the set of integers. The element $(0,\infty)$ will sometimes be written as 0, element $(+,0)$ as 1, and element $(-,0)$ as $-1$. Define function $\mathbb{R}^* \to \mathbb{R}^o$ by

$$r \mapsto r^o = \begin{cases} (\text{sign}(r), \hat{r}) & \text{if } r \neq 0 \\ 0 & \text{if } r = 0 \end{cases}$$

where $\text{sign}(r) = +$ if $r > 0$ and $\text{sign}(r) = -$ if $r < 0$.

The element $(+, n)$ will be used for reasoning about elements $r$ of $\mathbb{R}^*$ of positive sign and of order $\varepsilon^n$, i.e., $r$ such that $r^o = (+, n)$, for example $\varepsilon^n$ and $3\varepsilon^n \frac{1+\varepsilon}{1-\varepsilon}$. Similarly $(-, n)$ represents $r$ such that $r^o = (-, n)$, e.g., $-2\varepsilon^n$. Note that the operation $r \mapsto r^o$ is not onto; the image is the set $(\mathbb{R}^o \setminus \mathbb{R}^o_0) \cup \{0\}$ where $\mathbb{R}^o_0 = \{(0, n) : n \in Z \cup \{\infty\}\}$.

If $r^o = (+, n)$ and $s^o = (-, n)$ then $(r + s)^o$ could be 0, $(+, m)$ or $(-, m)$ for any $m \geq n$. For example, if $r = \varepsilon^n$ and $s = -2\varepsilon^n$ then $r^o = (+, n)$, $s^o = (-, n)$ and $(r + s)^o = (-, n)$; if, however, $s = -\varepsilon^n$ then $(r + s)^o = 0$. This is the reason that we included elements $(0, n)$ in $\mathbb{R}^o$: to ensure that $\mathbb{R}^o$ is closed under addition. $(0, n)$ is intended to represent elements $r$ of $\mathbb{R}^*$ with $\hat{r} \geq n$.

Thus elements of $\mathbb{R}^o$ are interpreted as representing certain subsets of $\mathbb{R}^*$. For $a \in \mathbb{R}^o$ we define subset $a^*$ of $\mathbb{R}^*$ which will be viewed as the 'meaning' of $a$. The calculus we will develop for $\mathbb{R}^o$ may be viewed as a simple way of reasoning about these subsets.

For $a \in \mathbb{R}^o \setminus \mathbb{R}^o_0$, let $a^* = \{r \in \mathbb{R}^* : r^o = a\}$. Thus, for $n \in Z$, $(+, n)^*$ is the set of positive elements of $\mathbb{R}^*$ of order $n$, $\{r : r > 0, \hat{r} = n\}$, and $(-, n)^*$ is the set of negative elements of $\mathbb{R}^*$ of order $n$, $\{r : r < 0, \hat{r} = n\}$. For $(0, m) \in \mathbb{R}^o_0$, let $(0, m)^* = \{r \in \mathbb{R}^* : \hat{r} \geq m\}$, so that, for example, $(0, \infty)^* = \{0\}$. Elements $a^*$ for $a \in (\mathbb{R}^o \setminus \mathbb{R}^o_0) \cup \{0\}$ form of partition of $\mathbb{R}^*$.

### 4.2 The Operations on $\mathbb{R}^o$

The operations on $\mathbb{R}^*$ induce operations on $\mathbb{R}^o$ (the definitions are formally justified by Theorem 1 in section 5.1). These are related to operations in Parsons' work on qualitative uncertainty e.g. [Parsons, 93].

**Multiplication:** For $(\sigma, m), (\sigma', n) \in \mathbb{R}^o$, let $(\sigma, m) \times (\sigma', n) = (\sigma \otimes \sigma', m + n)$, where $\infty + m = m + \infty = \infty$ for $m \in Z \cup \{\infty\}$, and $\otimes$ is the natural multiplication of signs: it is the commutative operation on $\{+,-,0\}$ such that $+ \otimes - = -$, $+ \otimes + = - \otimes - = +$, and for any $\sigma \in \{+,-,0\}$, $\sigma \otimes 0 = 0$. As usual, $a \times b$ will be sometimes abbreviated to $ab$. This multiplication is associative and commutative, and $(\mathbb{R}^o \setminus \mathbb{R}^o_0, \times)$ is an abelian group. Also $-1 \times -1 = 1$ and for any $a \in \mathbb{R}^o$, $a \times 0 = 0$ and $a \times 1 = a$.

For $b \in \mathbb{R}^o \setminus \mathbb{R}^o_0$ define $b^{-1}$ to be the multiplicative inverse of $b$, and for $a \in \mathbb{R}^o$ let $a/b = a \times b^{-1}$. $(\sigma, m)^{-1} = (\sigma, -m)$ for $\sigma \in \{+,-\}$.

**Addition:** For $(\sigma,m),(\sigma',n) \in \mathbb{R}^o$, let

$$(\sigma,m)+(\sigma',n) = \begin{cases} (\sigma,m) & \text{if } m < n; \\ (\sigma',n) & \text{if } m > n; \\ (\sigma \oplus \sigma',m) & \text{if } m = n \end{cases}$$

where $+\oplus+ = +$, $-\oplus- = -$, and otherwise, $\sigma \oplus \sigma' = 0$.

Addition is associative and commutative, and $a+0 = a$ for any $a \in \mathbb{R}^o$. We have distributivity: for $a,b,c \in \mathbb{R}^o$, $(a+b)c = ac+bc$.

For $a,b \in \mathbb{R}^o$ let $-b = -1 \times b$, and $a-b = a+(-b)$. We have $-(\sigma,m) = (-(\sigma),m)$ where, as one would expect, $-(+) = -$, $-(-) = +$ and $-(0) = 0$. If $a,b,c,d \in \mathbb{R}^o$ and $b,d \notin \mathbb{R}^o_0$ then $\frac{a}{b} = \frac{ad}{bd}$, $\frac{a}{b}+\frac{c}{b} = \frac{a+c}{b}$, $\frac{a}{b}+\frac{c}{d} = \frac{ad+bc}{bd}$, and $\frac{a}{b}-\frac{c}{d} = \frac{ad-bc}{bd}$. The above properties mean that arithmetic expressions in $\mathbb{R}^o$ can be manipulated in many of the ways that arithmetic expressions in $\mathbb{R}$ can; for example, we shall see in section 7.1 that order of magnitude expectation has the usual linearity properties. However $(\mathbb{R}^o,0,1,+,\times)$ is not a field, or even a ring, because additive inverses do not exist; for example, $1+-1 = (0,0)$ which is not equal to $0 = (0,\infty)$.

### 4.3 Ordering

Define $(\sigma,m) > 0$ iff $\sigma = +$, and $(\sigma,m) > (\sigma,n)$ iff $(\sigma,m)-(\sigma,n) > 0$.

We have that for $a,b \in \mathbb{R}^o$, $a > 0$ if and only if $0 > -a$, and $a,b > 0$ implies that $a+b > 0$ and $ab > 0$. $>$ is a transitive relation, but is not a total order since, for example, we have neither $(0,0) > -1$ nor $(0,0) < -1$. Unfortunately, it is possible that $a > b$ and $c > 0$ hold but $a+c > b+c$ does not hold. For example, if $a = 1, b = 0, c = -1$ then $a > b$ but $a+c = (0,0)$ which is not greater than $b+c = -1 = (-,0)$; the reason for this is that one of the elements that $(0,0)$ represents (e.g., $-2 \in \mathbb{R}^*$) is smaller than one of the elements that $(-,0)$ represents (e.g., $-1 \in \mathbb{R}^*$).

## 5 Interpretation of the Order of Magnitude Calculus

Here we give a precise semantics for the order of magnitude calculus. The meaning of elements (and hence subsets) of $\mathbb{R}^*$ is given above. As stated there, each element $a$ of $\mathbb{R}^o$ will be viewed as a representation of the set $a^* \subseteq \mathbb{R}^*$. Thus order of magnitude statements are interpreted as statements about $\mathbb{R}^*$. Extended real $r \in a^*$ will be described as ***an interpretation of*** $a$. Each $a \in \mathbb{R}^o$ is considered to be representing some unknown interpretation $r \in \mathbb{R}^*$.

We will show that the calculus is sound and complete in a particular sense. Roughly speaking, soundness will mean that any computation in the order of magnitude calculus is correct when viewed as a statement about subsets of $\mathbb{R}^*$; completeness will mean that the calculus is as strong as it could be. For example, suppose that for particular elements $a,b,c,d \in \mathbb{R}^o$, $(a+b)/c > c-d$ holds; we will show that for all interpretations $r$ of $a$, $s$ of $b$, $t$ and $u$ of $c$ and $v$ of $d$, $(r+s)/t > u-v$ (note that we should not assume that the two instances of $c$ in the first equation represent the same unknown element of $\mathbb{R}^*$). If this were not the case then the order of magnitude calculus would be making unwarranted conclusions. This is a soundness result for the order of magnitude calculus. Conversely, if for all interpretations $r$ of $a$, $s$ of $b$, $t$ and $u$ of $c$ and $v$ of $d$, $(r+s)/t > u-v$, then we have that $(a+b)/c > c-d$. If this were not the case then the order of magnitude calculus would be weaker than it ideally should be. This is a completeness result.

### 5.1 Representation Within Algebra of Subsets of $\mathbb{R}^*$

For $S,T \subseteq \mathbb{R}^*$ and $U \subseteq \mathbb{R}^* \setminus \{0\}$, define

$$\begin{aligned} S+T &= \{s+t : s \in S, t \in T\} \\ ST &= \{st : s \in S, t \in T\} \\ -S &= \{-s : s \in S\} \\ U^{-1} &= \{u^{-1} : u \in U\} \end{aligned}$$

Define relation $>$ on subsets of $\mathbb{R}^*$ by $S > T$ if and only if for all $s \in S$ and $t \in T$, $s > t$.

($ST$ will sometimes be (implicitly) referred to as $S \times T$, although this notation is used as little as possible to avoid confusion with product sets.) The following result justifies the operations in the order of magnitude calculus.

**Theorem 1**

For $a,b,c \in \mathbb{R}^o$ with $c \notin \mathbb{R}^o_0$,

(i) $a \in \mathbb{R}^o_0 \iff a^* \ni 0$;

(ii) $(a+b)^* = a^*+b^*$, $(ab)^* = a^*b^*$, $(-a)^* = -(a^*)$, $(c^{-1})^* = (c^*)^{-1}$;

(iii) $a = b \iff a^* = b^*$, and $a > b \iff a^* > b^*$.

In other words, the structures $(\mathbb{R}^o,+,\times,>)$ and $(\{a^* : a \in \mathbb{R}^o\},+,\times,>)$ are isomorphic, with isomorphism $a \mapsto a^*$.

The proof is tedious but straight-forward. One might imagine that all the important properties of the order of magnitude calculus can be derived from those of subsets of $\mathbb{R}^*$. This is not the case; we are fortunate that the very valuable distributivity property holds in the order of magnitude calculus, and hence in the calculus of the corresponding subsets of $\mathbb{R}^*$, since it does not hold for arbitrary subsets of $\mathbb{R}^*$; for example, if $S = T = \{1\}$, $U = \{1,2\}$, then $(S+T)U = \{2\}U = \{2,4\}$, but $SU+TU = U+U = \{2,3,4\}$.

### 5.2 Completeness Results for Arithmetic Statements

We will construct arithmetic formulae based on set of symbols $X = \{x_1, x_2, \ldots\}$. Define the set of arithmetic formulae $\mathcal{A}$ to be the smallest set of strings of symbols such that

(i) for all $i = 1, 2, \ldots$, the symbol $x_i$ is in $\mathcal{A}$; $x_i$ is said to have symbol set $\{x_i\}$;

(ii) if $\varphi, \psi \in \mathcal{A}$ with symbol sets $Y$ and $Z$ respectively, and $Y \cap Z = \emptyset$, then the following (strings of symbols) are in $\mathcal{A}$: $(-\varphi)$ and $(\varphi^{-1})$, which both have symbol set $Y$, $(\varphi + \psi)$ and $(\varphi \times \psi)$ which both have symbol set $Y \cup Z$.

For example, $\big(x_2 + (-(x_1 \times x_3))\big)$ is an arithmetic formula with symbol set $\{x_1, x_2, x_3\}$. Note that each symbol $x_i$ can appear at most once in an arithmetic formula.

Consider some tuple $R = (R, R', +, \times, -(\cdot), (\cdot)^{-1})$, where $R$ is a set, $R' \subseteq R$, $+$ and $\times$ are binary operations on $R$, $-(\cdot)$ is a function from $R$ to $R$, and $(\cdot)^{-1}$ is a function from $R'$ to $R$.

An ***instantiation*** $\underline{a}$ ***in*** $R$ ***for arithmetic formula*** $\varphi$ is a function from $Y$ to $R$, where $Z \subseteq Y \subseteq X$ and $Z$ is the symbol set of $\varphi$. $\varphi(\underline{a})$ is then defined to be the value of the arithmetic expression $\varphi$ when symbol $x_i$ is replaced by $\underline{a}(x_i)$ for each $x_i \in Z$. Formally it is defined inductively as follows: $x_i(\underline{a}) = \underline{a}(x_i)$, $(\varphi + \psi)(\underline{a}) = \varphi(\underline{a}) + \psi(\underline{a})$, $(\varphi \times \psi)(\underline{a}) = \varphi(\underline{a}) \times \psi(\underline{a})$, $(-\varphi)(\underline{a}) = -\big(\varphi(\underline{a})\big)$, and $(\varphi^{-1})(\underline{a}) = \big(\varphi(\underline{a})\big)^{-1}$ if $\varphi(\underline{a}) \in R'$, and $(\varphi^{-1})(\underline{a})$ equals the string 'undefined' otherwise.

The definition allows us to instantiate different symbols $x_i$ and $x_j$ with the same value $r \in R$. For example, if $\varphi = \big(x_2 + (-(x_1 \times x_3))\big)$, $R = I\!R^o$, and $\underline{a}(x_1) = \underline{a}(x_2) = (-, 2)$, $\underline{a}(x_3) = (+, 0)\ (= 1)$, then $\varphi(\underline{a}) = (-, 2) + \big(-((-, 2) \times 1)\big) = (-, 2) + (+, 2) = (0, 2)$.

For arithmetic formula $\varphi$ and instantiation $\underline{a} : Y \to I\!R^o$ we say that $\underline{r} : Y \to I\!R^*$ is an ***interpretation of*** $\underline{a}$ (sometimes abbreviated to '$\underline{r}$ is of $\underline{a}$') if for all $x_i \in Y$, $\underline{r}(x_i) \in \big(\underline{a}(x_i)\big)^*$, i.e., if each component of $\underline{r}$ is an interpretation of each component of $\underline{a}$. Clearly, $\underline{r}$ is an instantiation in $I\!R^*$ for $\varphi$. Also define instantiation $\underline{a}^*$ in $2^{I\!R^*}$ by $\underline{a}^*(x_i) = \big(\underline{a}(x_i)\big)^*$ for each $x_i \in Y$; naturally, for $R = I\!R^o$, we define $R'$ to be $I\!R^o \setminus I\!R^o_0$, for $R = I\!R^*$, $R' = I\!R^* \setminus \{0\}$, and for $R = 2^{I\!R^*}$, $R' = \{U : U \subseteq I\!R^* \setminus \{0\}\}$.

**Theorem 2**

Let $\underline{a}$ and $\underline{b}$ be instantiations in $I\!R^o$ for arithmetic formulae $\varphi$ and $\psi$, respectively. Then

(i) $\varphi(\underline{a}) = \text{undefined} \iff \varphi(\underline{a}^*) = \text{undefined} \iff$ for some interpretation $\underline{r}$ of $\underline{a}$, $\varphi(\underline{r}) = \text{undefined}$;

(ii) when $\varphi(\underline{a}) \neq \text{undefined}$ the following equivalences hold: $\varphi(\underline{a}) = \psi(\underline{b}) \iff \varphi(\underline{a}^*) = \psi(\underline{b}^*) \iff \{\varphi(\underline{r}) : \underline{r} \text{ is of } \underline{a}\} = \{\psi(\underline{s}) : \underline{s} \text{ is of } \underline{b}\}$;

(iii) $\varphi(\underline{a}) > \psi(\underline{b}) \iff \varphi(\underline{a}^*) > \psi(\underline{b}^*) \iff$ for all interpretations $\underline{r}$ of $\underline{a}$, and $\underline{s}$ of $\underline{b}$, $\varphi(\underline{r}) > \psi(\underline{s})$.

(iii) can be viewed as a soundness and completeness result for the order of magnitude calculus; any strict inequality statement that can be deduced in the order of magnitude calculus is true when viewed as a statement about the extended reals (soundness); conversely, any strict inequality statement in the order of magnitude calculus which is true when viewed as a statement in the extended reals, can be deduced in the order of magnitude calculus (completeness).

## 6 Order of Magnitude Probability/Utility

To define order of magnitude probability and utility, we use, as before, the standard definitions, except using $I\!R^o$ instead of $I\!R$.

An order of magnitude utility function on $\Omega$ is a function from $\Omega$ to $I\!R^o$.

An order of magnitude probability function P over $\Omega$ is defined to be an $I\!R^o$-valued probability function (with $R' = I\!R^o \setminus I\!R^o_0$; see definition of $R$-valued probability function in section 2.3).

### 6.1 The Meaning of Order of Magnitude Probability

If $f$ is a function from some set $W$ to $I\!R^o$ then $g: W \to I\!R^*$ is said to be ***an interpretation of*** $f$ if for all $w \in W$, $g(w) \in \big(f(w)\big)^*$. If $g$ happens also to be an extended probability function then we say that ***$g$ is a probabilistic interpretation of $f$***.

Each order of magnitude probability function P over $\Omega$ has a probabilistic interpretation; for example we can define extended probability function $R$ by $R(\omega) = 0$ if $\mathrm{P}(\omega) = 0$ and $R(\omega) = \alpha\epsilon^n$, for $\mathrm{P}(\omega) = (+, n)$, where $\alpha$ is a normalisation constant (and then we extend $R$ by additivity to $2^\Omega$). Conversely, each extended probability function $R$ is a probabilistic interpretation for exactly one order of magnitude probability function $R^o$ given by $R^o(A) = \big(R(A)\big)^o$; also for all $A, B \subseteq \Omega$ with $R^o(B) \notin I\!R^o_0$, $R^o(A|B) = \big(R(A|B)\big)^o$. Similar comments apply for utility.

This means that an order of magnitude probability function P may be viewed as a representation of a set of extended probability functions, i.e., its probabilistic interpretations, which were given meaning in section 3 and in [Wilson, 95].

Thus if $\mathrm{P}(A) = (+, 1)$ then we judge the probability of $A$ to be of the same order as the calibrating event $E$ (see section 3.3); if $\mathrm{P}(B) = (+, n)$ we judge the probability of $B$ be of the same order as $n$ independent events, each of which has probability equal to that of $E$.

### 6.2 Kappa Functions

A kappa function over $\Omega$ is defined to be a function $\kappa: 2^{\Omega} \to I\!N \cup \{\infty\}$ such that

(i) $\kappa(\emptyset) = \infty$; $\kappa(\Omega) = 0$;

(ii) for $A, B \subseteq \Omega$ such that $A \cap B = \emptyset$, $\kappa(A \cup B) = \min(\kappa(A), \kappa(B))$.

For kappa function $\kappa$ over $\Omega$ and $A, B \subseteq \Omega$ such that $\kappa(B) \neq \infty$, the conditional value $\kappa(A|B)$ is defined to be $\kappa(A \cap B) - \kappa(B)$.

Kappa functions are a slight generalisation of Spohn's NCFs [Spohn, 88, 90], in that a non-empty set $A$ can be assigned value $\infty$, meaning '$A$ is impossible'; they have been used, for example, in [Goldszmidt and Pearl, 91; Goldszmidt, 92; Pearl, 93a,b]. They are also closely related to Zadeh's possibility functions [Dubois and Prade, 88]; the mapping $\kappa \mapsto 2^{-\kappa}$ gives an embedding of kappa functions into the set of possibility measures.

The values of an order of magnitude probability function are all contained in $[0,1]^{\circ} = \{(+, m) : m \geq 0\} \cup \{0\}$. This set is totally ordered by $>$, and in fact $([0,1]^{\circ}, +, \times, >)$ can be seen to be isomorphic to $(I\!N \cup \{\infty\}, \min, +, <)$. Using this isomorphism, it is easy to see that order of magnitude probability functions are just kappa functions, with their values labelled differently; also definitions of conditional values $\kappa(A|B)$ and $P(A|B)$ can be seen to be equivalent. This means that the justification above for order of magnitude probabilities also justifies kappa functions. Of course, this justification is closely related to standard justifications in terms of infinitesimal probabilities such as [Spohn, 90], but I think that it is stronger since it is based on a more meaningful theory of infinitesimal probability (see section 3). The justification also benefits logics such as Adams' [66,75] which can be given semantics in terms of kappa functions (see e.g., [Goldszmidt, 92]).

## 7 Order of Magnitude Decision Theory

This section shows how the definitions for order of magnitude probability and utility lead to an order of magnitude decision theory, and the relationship with the decision theory for kappa functions in [Pearl, 93a] is discussed.

### 7.1 Expectation

Expectation for order of magnitude probability functions can be defined in the usual way, and turns out to have the usual linearity properties.

**Proposition**

Let P be an order of magnitude probability function over $\Omega$, and let $U$ and $V$ be order of magnitude utility functions on $\Omega$. Define $P(U)$, the expected value of $U$ with respect to P, to be $\sum_{\omega \in \Omega} P(\omega)U(\omega)$ (note that the summation sign makes sense because of the commutativity and associativity of addition in $I\!R^{\circ}$). For $\lambda \in I\!R^{\circ}$, let $U_{\lambda}$ be the constant function on $\Omega$ which takes value $\lambda$. Then,

(i) $P(U + V) = P(U) + P(V)$;

(ii) $P(-U) = -P(U)$;

(iii) $P(U_{\lambda}) = \lambda$,

(iv) $P(\lambda U) = \lambda P(U)$,

where, as usual, addition and scalar multiplication of the order of magnitude utility functions is defined pointwise.

This result follows easily from the properties of the order of magnitude calculus given in section 4.2.

Let P and $U$ be an order of magnitude probability and utility function on $\Omega$, respectively. Let $\Omega' = \{\omega \in \Omega : P(\omega) \neq 0\}$. For $\omega \in \Omega'$ let us write $P(\omega)$ as $(+, \kappa(\omega))$ and $U(\omega)$ as $(\sigma(\omega), m(\omega))$. Then, $P(U)$ equals

$$\sum_{\omega \in \Omega} P(\omega)U(\omega) = \sum_{\omega \in \Omega'} (+, \kappa(\omega))\,(\sigma(\omega), m(\omega)),$$

which equals $(+, u^+) + (0, u^0) + (-, u^-) = (\sigma, u)$ where, for $\chi \in \{+, -, 0\}$, $u^{\chi} = \min_{\omega \in \Omega' : \sigma(\omega) = \chi} (\kappa(\omega) + m(\omega))$ (the operator min taken over the empty set is defined to have value $\infty$), $u = \min(u^+, u^0, u^-)$, and

$$\sigma = \begin{cases} + & \text{if } u^+ < u^0, u^-; \\ - & \text{if } u^- < u^0, u^+; \\ 0 & \text{otherwise.} \end{cases}$$

### 7.2 Comparison of Expected Utility

Suppose we are comparing two options with associated order of magnitude probability and utility functions $P_i, U_i$ for $i = 1, 2$. Option 1 is strictly preferred to option 2 if $P_1(U_1) > P_2(U_2)$.

**Theorem 3**

For $i = 1, 2$, let $P_i$ be order of magnitude probability functions over $\Omega$ and $U_i$ be order of magnitude utility functions on $\Omega$. Then $P_1(U_1) > P_2(U_2)$ if and only if for any probabilistic interpretations $R_i$ of $P_i$ ($i = 1, 2$), and for any interpretations $V_i$ of $U_i$ ($i = 1, 2$), $R_1(V_1) > R_2(V_2)$.

This can be viewed as a soundness and completeness result for the order of magnitude decision theory: we strictly prefer option 1 to option 2 if and only if for all probabilistic interpretations we would do so.

### 7.3 Pearl's Decision Theory for Kappa Functions

In [Pearl, 93a, section 3] a decision theory for kappa functions is developed.[2] The scale used for utility is coarser than that given here, and leads sometimes to conclusions being unnecessarily weak.

Pearl's order of magnitude utility functions on $\Omega$ are functions $\mu$ from $\Omega$ to $Z$, the set of integers. For $i > 0$, $\mu(\omega) = i$ is intended to represent values of utility of order $1/\epsilon^i$, and thus corresponds to $(+, -i) \in \mathbb{R}^o$; $\mu(\omega) = -i$ is intended to represent values of utility of order $-1/\epsilon^i$, thus corresponds to $(-, -i) \in \mathbb{R}^o$. It seems that $\mu(\omega) = 0$ is intended to represent values of utility of order $+1$ or $-1$ or anything in between (i.e., constant function of $\epsilon$ or less), so it corresponds to $(0,0) \in \mathbb{R}^o$.

Note that, because of the reasons given earlier, in the discussion on addition in $\mathbb{R}^o$, the sum of $\mu$ values $i(> 0)$ and $-i$ does not correspond to a single $\mu$ value, so when such a value occurs in a calculation, it is labelled 'ambiguous'. Thus elements $(0, -i) \in \mathbb{R}^o$ for $i > 0$ are represented in Pearl's system by the value of 'ambiguous'.

Let $\kappa$ be a kappa function over $\Omega$. For $i = 0, 1, 2, \ldots$, the set $W_i^+$ is defined to be $\{\omega \in \Omega : \mu(\omega) = i\}$ and $W_i^-$ is defined to be $\{\omega \in \Omega : \mu(\omega) = -i\}$. Define non-negative integers $n^+$ and $n^-$ by $n^+ = \max_i(0, i - \kappa(W_i^+))$ and $n^- = \max_i(0, i - \kappa(W_i^-))$.

The expected utility of $\mu$ with respect to $\kappa$ is then defined in [Pearl, 93a] to be

$$\begin{cases} \text{ambiguous} & \text{if } n^+ = n^- > 0; \\ n^+ - n^- & \text{otherwise.} \end{cases}$$

As we shall shortly see, this cannot have been what Pearl intended; instead he probably intended the expected utility $e$ of $\mu$ with respect to $\kappa$ to be

$$\begin{cases} \text{ambiguous} & \text{if } n^+ = n^- > 0; \\ 0 & \text{if } n^+ = n^- = 0; \\ n^+ & \text{if } n^+ > n^-; \\ -n^- & \text{if } n^+ < n^-. \end{cases}$$

#### Example

Let $\Omega = \{\omega_1, \omega_2\}$, and define $\kappa$ by $\kappa(\omega_1) = \kappa(\omega_2) = 0$, corresponding to non-infinitesimal values of probability, and $\mu$ by $\mu(\omega_1) = 4$, $\mu(\omega_2) = -3$, which correspond to utilities of orders $1/\epsilon^4$ and $-1/\epsilon^3$ respectively. Thus the expected utility will be of order $1/\epsilon^4 - 1/\epsilon^3$ which is of order $1/\epsilon^4$. We have $n^+ = 4$, $n^- = 3$, so the original definition of expected value of $\mu$ with respect to $\kappa$ gives a value of 1 corresponding to expected utility of order $1/\epsilon$, not $1/\epsilon^4$.

[2] Pearl refers to this as a qualitative decision theory, and the kappa function as an ordinal belief ranking; this is misleading: an ordinal scale is not sufficient for kappa functions because the differences between values do matter, for conditioning and independence.

The second definition gives the correct answer in this example and will be justified using the order of magnitude calculus.

As explained earlier, kappa function $\kappa$ corresponds to an order of magnitude probability function P, and, using the above correspondence, $\mu$ functions can be viewed as order of magnitude utility functions $U$; for example $\mu(\omega) = -5$ is translated to $U(\omega) = (-, -5)$ and $\mu(\omega) = 0$ is translated to $U(\omega) = (0,0)$. Thus we can use the results of 7.1 to give the correct expected value. Using the results and notation of section 7.1 we have $\mathrm{P}(U) = (\sigma, u)$ where $\sigma$ equals

$$\begin{cases} 0 & \text{if } u = u^0 \\ 0 & \text{if } u \neq u^0 \text{ and } u^+ = u^-; \\ + & \text{if } u \neq u^0 \text{ and } u^+ < u^-; \\ - & \text{if } u \neq u^0 \text{ and } u^+ > u^-. \end{cases}$$

By considering some $\omega' \in \Omega$ with $\kappa(\omega') = 0$, we can see that, because of the special form of $U$, (i) $u \leq 0$ and (ii) $u = u^0 \iff u = 0 \iff u^+, u^- \geq 0$. These facts enable us to eliminate mentions of $u^0$ and $u$ in the above equation. Proving that $n^+ = \max(0, -u^+)$ and $n^- = \max(0, -u^-)$ is straight-forward, and these equivalences can be used to show that $\mathrm{P}(U)$ equals

$$\begin{cases} (0,0) & \text{if } n^+ = n^- = 0 \\ (0, -n^+) & \text{if } n^+ = n^- > 0; \\ (+, -n^+) & \text{if } n^+ > n^-; \\ (-, -n^-) & \text{if } n^+ < n^-. \end{cases}$$

When we convert the order of magnitude values into Pearl's $\mu$-values we get the amended definition.

This justifies Pearl's (amended) decision theory. Note, that we have just shown a soundness result; it does sometimes give unnecessarily weak results, as shown by the following example, in which we have a choice between two options with associated $\kappa$ and $\mu$ functions. Let $\Omega = \{\omega_1, \omega_2\}$. For $i = 1, 2$, let $\kappa_i(\omega_1) = \kappa_i(\omega_2) = 0$. Let $\mu_1(\omega_1) = 2$ and $\mu_2(\omega_2) = -2$, and let $\mu_2(\omega_1) = -5 = \mu_2(\omega_2)$. The expected value $e_1$ of $\mu_1$ with respect to $\kappa_1$ equals 'ambiguous', and $e_2 = -5$ so neither option is preferred over the other. However, the expected utility in the first option is of order $\pm 1/\epsilon^2$ which is greater than the expected utility of the second option, of order $-1/\epsilon^5$. The order of magnitude calculus is able to give this conclusion (as we know from the completeness results): $\mathrm{P}_1(U_1) = (0, -2) > (-, -5) = \mathrm{P}_2(U_2)$.

Thus, although we have given Pearl's order of magnitude decision theory a formal semantics, it has a number of disadvantages compared to one developed in this paper; it cannot represent infinitesimal utilities, it does not distinguish utilities of order 1, representing finite positive benefit, from utilities of order $-1$, representing finite negative benefit; also, it does not distinguish different grades of ambiguity, which means that it lacks the completeness results enjoyed by this paper's formalism.

## 8 Discussion

This is clearly a rather simple order of magnitude calculus; it was designed to be just expressive enough to give a satisfactory decision theory for kappa functions. There are natural ways to extend the calculus; we might add elements $(p, n)$ representing the non-negative elements of order at most $n$, and $(m, n)$ representing the non-positive elements of order at most $n$, so that e.g., $(p, n)^* = \{r \in \mathbb{R}^* : \hat{r} \geq n, r \geq 0\}$. We might also then consider adding a squaring operation. The semantics of this would be given by defining for $S \subseteq \mathbb{R}^*$, $S^2 = \{s^2 : s \in S\}$. (note that $S^2$ is not usually equal to $SS$).

More ambitious extensions might allow representations of statements such as 'greater than order $\epsilon^n$', or such as 'between order $\epsilon^n$ and $\epsilon^m$'. We could also consider intermediate calculi; a natural idea is to define a calculus containing pairs $(\lambda, n)$, where $\lambda \in \mathbb{R} \setminus \{0\}$, $n \in Z$, representing extended reals $r$ with $\hat{r} = n$ and $\bar{r} = \lambda$; however, it may be the case that we would have to either give up distributivity or completeness. In any case, the methods and concepts of this paper, particularly section 5, would be useful in the analysis of these more sophisticated calculi.

It would be interesting to explore applications of this calculus in other areas of Artificial Intelligence, in particular, Qualitative Physics and Constraint-based reasoning.

### Acknowledgements

This work was supported by European Community ESPRIT project DRUMS2, BRA 6156. I benefitted from discussions with Moisés Goldszmidt, Milan Studený and František Matúš and technical assistance from David Shrimpton.